\documentclass[letterpaper, 10 pt, conference]{ieeeconf}  % Comment this line out if you need a4paper

\IEEEoverridecommandlockouts                              % This command is only needed if 
\usepackage{graphicx}
\usepackage{colortbl}

\usepackage{times}
\usepackage{epsfig}
\usepackage{amsmath}
\usepackage{amssymb}
\usepackage{arydshln}
\usepackage{color}
\usepackage{caption}
\usepackage{url}
\usepackage{here}
\usepackage{cite}

\usepackage[ruled,vlined]{algorithm2e}
\newcommand{\colornewparts}[0]{\color{black}}

\renewcommand{\baselinestretch}{0.98}

\usepackage{multirow,array}

\title{\LARGE \bf
Depth360: Self-supervised Learning for Monocular Depth Estimation using Learnable Camera Distortion Model
}

\author{Noriaki Hirose$^{1}$ and Kosuke Tahara$^{1}$% <-this % stops a space
\thanks{$^{1}$Noriaki Hirose and Kosuke Tahara are in Toyota Central R\&D Labs., Inc., 41-1, Yokomichi, Nagakute, Aichi, 480-1192, Japan
        {\tt\small hirose@mosk.tytlabs.co.jp}}%
}

\begin{document}

\maketitle
\thispagestyle{empty}
\pagestyle{empty}

%%%%%%%%%%%%%%%%%%%%%%%%%%%%%%%%%%%%%%%%%%%%%%%%%%%%%%%%%%%%%%%%%%%%%%%%%%%%%%%%
\begin{abstract}
Self-supervised monocular depth estimation has been widely investigated to estimate depth images and relative poses from RGB images. This framework is promising because the depth and pose networks can be trained from just time-sequence images without the need for the ground truth depth and poses. 

In this work, we estimate the depth around a robot (360$^\circ$ view) using time-sequence spherical camera images, from a camera whose parameters are unknown. We propose a learnable axisymmetric camera model which accepts distorted spherical camera images with two fisheye camera images as well as pinhole camera images. In addition, we trained our models with a photo-realistic simulator to generate ground truth depth images to provide supervision. Moreover, we introduced loss functions to provide floor constraints to reduce artifacts that can result from reflective floor surfaces. We demonstrate the efficacy of our method using the spherical camera images from the GO Stanford dataset and pinhole camera images from the KITTI dataset to compare our method’s performance with that of baseline method in learning the camera parameters.
\end{abstract}
%%%%%%%%%%%%%%%%%%%%%%%%%%%%%%%%%%%%%%%%%%%%%%%%%%%%%%%%%%%%%%%%%%%%%%%%%%%%%%%%
%
\section{INTRODUCTION}
Accurately estimating three-dimensional (3D) information of objects and structures in the environment is essential for autonomous navigation and manipulation~\cite{thrun2002probabilistic,biswas2012depth}. Self-supervised monocular depth estimation without ground truth~(GT) depth images is one of the most popular approaches for obtaining 3D information~\cite{zhou2017unsupervised,casser2019depth,mahjourian2018unsupervised,yang2018every,hirose2021plg}. 
In self-supervised monocular depth estimation, there are several limitations including that this method requires the camera parameters, it cannot estimate the real scale of the depth image, and it functions poorly for highly reflective objects. {\colornewparts These limitations hinder the amount of available datasets for training and critical artifacts for robotics applications. }
%In self-supervised monocular depth estimation, there are several limitations including that this method requires the camera parameters, there is no scaling of the estimated depth image, and it functions poorly for highly reflective objects. {\colornewparts These limitations hinder the amount of available datasets for training and critical artifacts for robotics applications. }

This paper proposes a novel self-supervised monocular depth estimation approach for a spherical camera image view. We propose a learnable axisymmetric camera model to handle images from a camera whose camera parameter is unknown. Because this camera model is applicable to highly distorted images, such as spherical camera images based on two fisheye images, we can obtain 360$^\circ$ 3D point clouds all around a robot from only one camera. 

In addition to self-supervised learning with real images, we rendered many pairs of spherical RGB images and their corresponding depth images from a photo-realistic robot simulator. In training, we mixed these rendered images with real images to achieve sim2real transfer in an attempt to provide scaling for the estimated depth.
Moreover, we introduce additional cost functions to improve the accuracy of depth estimation for reflective floor areas. 
We provide supervision for estimated depth images from the future and past robot footprints, which are obtained from the reference velocities in the data collection. 
Main contributions are summarized as:
\begin{figure}[t]
  \begin{center}
      \includegraphics[width=0.95\hsize]{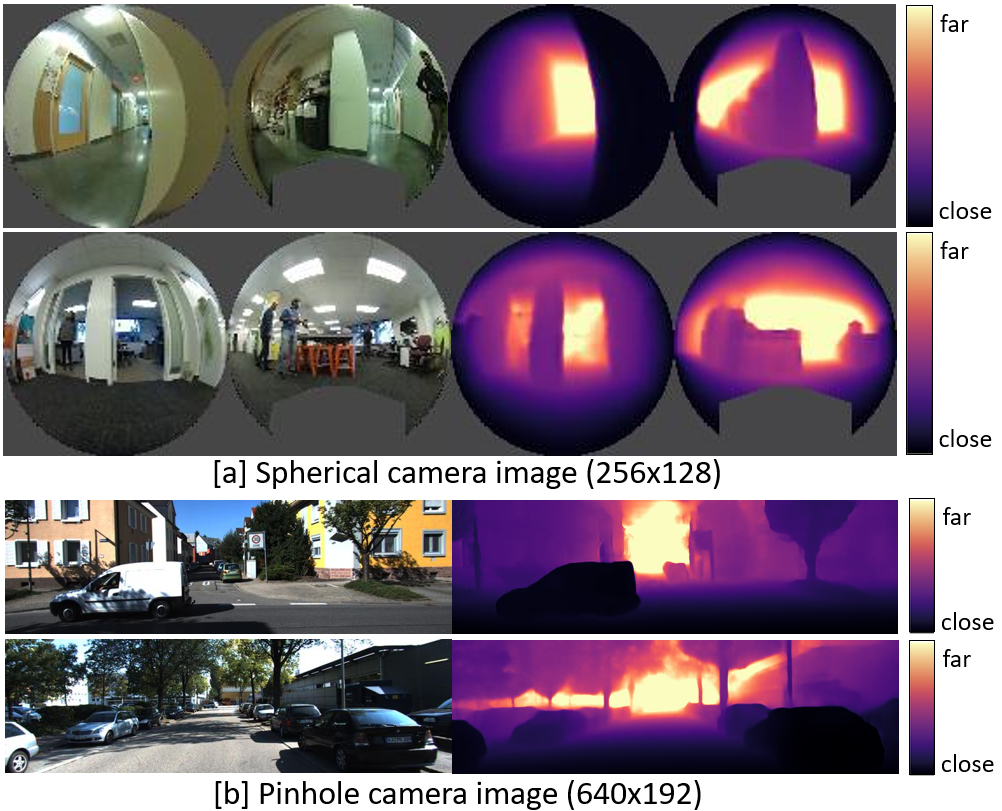}
  \end{center}
      \vspace*{-3mm}
	\caption{\small {\bf Estimated depth images from 256$\times$128 spherical camera images from the GO Stanford dataset and 640$\times$128 pinhole camera images from the KITTI dataset.} A spherical image was constructed using two fisheye images. The gray area in the spherical image is a common mask to exclude the corresponding pixels for training. Our method can estimate depth from the spherical image in [a] and the pinhole camera image in [b] using our learnable axisymmetric camera model.}
  \label{f:pull}
  \vspace*{-5mm}
\end{figure}
\begin{figure*}[ht]
  \vspace{1mm}
  \begin{center}
      \includegraphics[width=0.92\hsize]{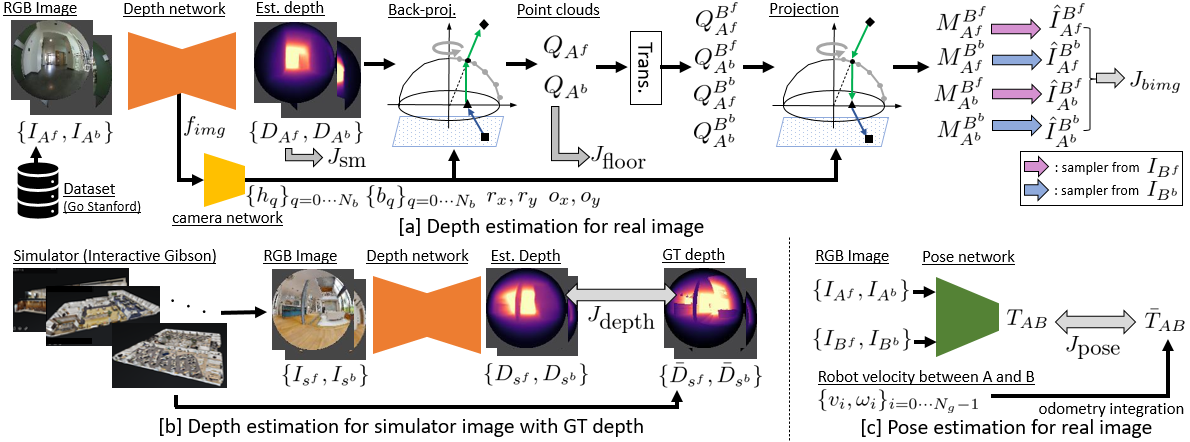}
  \end{center}
      \vspace*{-3mm}
	\caption{\small {\bf Block diagram of our method.} [a] Depth estimation for the real image of the GO Stanford dataset, [b] depth estimation for the simulation image with GT depth. The RGB image and GT depth are rendered via a 3D reconstructed environment using a Matterport scanner~\cite{matterport}, [c] pose estimation for the ``Trans.'' in [a].}
  \label{f:block}
  \vspace*{-5mm}
\end{figure*}
\vspace{-4mm}
\begin{itemize}
  \item A novel learnable axisymmetric camera model capable of handling distorted images with unknown camera parameters,
  \item Sim2real transfer using ground truth depth from a photo-realistic simulator to sharpen the estimated depth image and provide scaling,
  \item Proposal of novel loss functions that use the robot footprints and trajectories to provide constraints against reflective floor surfaces.
\end{itemize}
\vspace{0mm}
In addition to these main contributions, we blended front- and back-side fisheye images to reduce the occluded area for image prediction in self-supervised learning. As a result, our method can estimate the depth image without the large artifacts from a low-resolution spherical image. 

Our method was trained and evaluated on the GO Stanford~(GS) dataset~\cite{hirose2018gonet,hirose2019vunet,hirose2019deep}, as depicted in in Fig~\ref{f:pull}[a], with time-sequence spherical camera images and associated reference velocities for dataset collection. 
{\colornewparts Moreover, we solely evaluated the most important contribution, that is, the learnable axisymmetric camera model on the common KITTI dataset~\cite{Geiger2013IJRR}}, as depicted in Fig~\ref{f:pull}[b], to provide a comparison with other learnable camera models~\cite{gordon2019depth,vasiljevic2020neural}. 
We quantitatively and qualitatively evaluated our method with the GS and KITTI datasets.
\section{RELATED WORK}
Monocular depth estimation using self-supervised learning has been widely investigated by several approaches that use deep learning~\cite{garg2016unsupervised,godard2017unsupervised,guizilini2019packnet,yang2018every,wang2018learning,chen2019self,kumar2020unrectdepthnet,casser2019depth,gordon2019depth,guizilini2020robust,Poggi_CVPR_2020}. 
Zhou et al.~\cite{zhou2017unsupervised}, and Vijayanarasimhan et al.~\cite{vijayanarasimhan2017sfm} applied spatial transformer modules \cite{jaderberg2015spatial} based on the knowledge of structure from motion to achieve self-supervised learning from time-sequence images. 
Since the publication of \cite{zhou2017unsupervised} and \cite{vijayanarasimhan2017sfm}, several subsequent studies have attempted to estimate accurate depth using different methods, including a modified loss function~\cite{godard2019digging}, an additional loss function to penalize geometric inconsistencies~\cite{mahjourian2018unsupervised,zou2018df,gordon2019depth,luo2020consistent,hirose2021plg}, a probabilistic approach to estimate the reliability of depth estimation~\cite{Poggi_CVPR_2020,hirose2020variational}, masking dynamic objects~\cite{casser2019depth,gordon2019depth}, and an entirely novel model structure~\cite{guizilini2019packnet,yang2021transformer,lee2019big}.
We review the two categories most related to our method.

\noindent
\textbf{Fisheye camera image.} Various approaches have attempted to estimate depth images from fisheye images. \cite{kumar2018monocular} proposed supervised learning with sparse GT from LiDAR. \cite{won2019sweepnet,komatsu2020360, cui2019real} leveraged multiple fisheye cameras to estimate a  360$^\circ$ depth image. Similarly, Kumar et al. proposed self-supervised learning approaches with a calibrated fisheye camera model~\cite{kumar2020fisheyedistancenet} and semantic segmentation~\cite{kumar2021syndistnet}.

\noindent
\textbf{Learning camera model.} Gordon et al.~\cite{gordon2019depth} proposed a self-supervised monocular depth estimation method that can learn the camera's intrinsic parameters. Vasiljevic et al.~\cite{vasiljevic2020neural} proposed a learnable neural ray surface~(NRS) camera model for highly distorted images.

In contrast to the most related work~\cite{vasiljevic2020neural}, our learnable axisymmetric camera model is fully differentiable without the softmax approximation. Hence, end-to-end learning can be achieved without adjusting the hyperparameters during training. 
In addition, we provide supervision for the estimated depth images by using a photo-realistic simulator and robot trajectories from the dataset. As a result, our method can accurately estimate the depth from low-resolution spherical images.
\section{PROPOSED METHOD}
From the process in Fig.~\ref{f:block}, we designed the following cost functions to train the depth, pose and camera networks:
\begin{eqnarray}
    J = J_{bimg} + \lambda_{d} J_{depth} + \lambda_{f} J_{floor} + \lambda_{p} J_{pose} + \lambda_{s} J_{sm}.
    \label{eq:jall}
\end{eqnarray}

{\colornewparts In $J_{bimg}$, we propose a learnable camera model to handle the image sequences with unknown camera parameters. Our camera model can be trained without the GT camera parameters, through minimization of $J$ for self-supervised monocular depth estimation. The camera model has the learnable convex projection surface to deal with the arbitrary distortion, e.g. spherical camera image with two fisheye images. Contrary to the image loss of baseline methods~\cite{zhou2017unsupervised,godard2019digging}, $J_{bimg}$ is an occlusion-aware image loss using our learnable camera model to leverage the advantage of the 360$^\circ$ view around the robot by the spherical camera.}

$J_{depth}$ penalizes the depth difference using the GT depth from photo-realistic simulator. By penalizing $J_{depth}$ with $J_{bimg}$, our model can learn sim2real transfer and can thereby estimate accurate depths from real images. $J_{floor}$ is the proposed loss function that provides supervision against floor areas by using robot's footprint and trajectory. 

{\colornewparts In addition to these major contributions}, $J_{pose}$ penalizes the difference between the estimated pose and the GT pose calculated from the reference velocities in the dataset. $J_{sm}$ penalizes the discontinuity of the estimated depth using the exactly same objective, following \cite{godard2017unsupervised,godard2019digging}.

In the following sections, we first present $J_{bimg}$ and $J_{pose}$ in the overview of the training process. Then, we explain our camera model as the main contribution. Finally, we introduce $J_{depth}$ and $J_{floor}$ to improve the performance of depth estimation.

\begin{figure}[t]
  \vspace{1.5mm}
  \begin{center}
      \includegraphics[width=0.9\hsize]{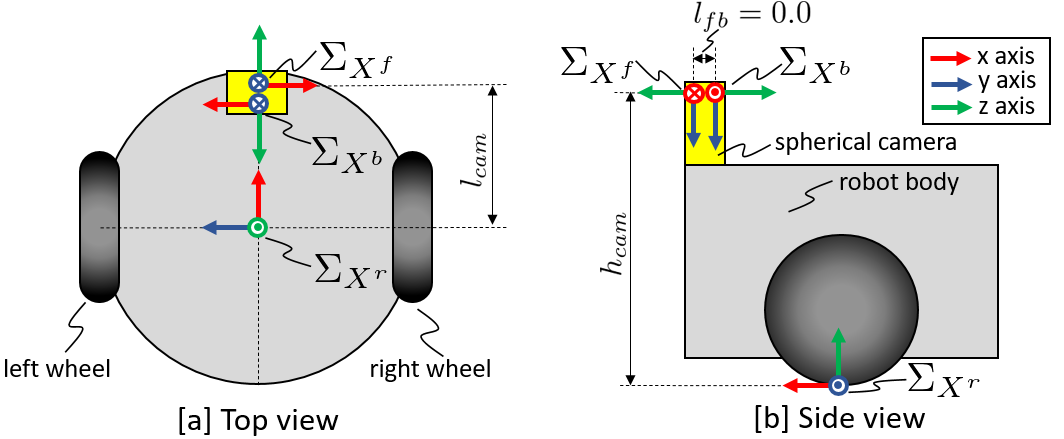}
  \end{center}
      \vspace*{-3mm}
	\caption{\small {\bf Robot and spherical camera coordinates.} $\Sigma_{X^f}$ and $\Sigma_{X^b}$ are the coordinates of the front- and back-side fisheye cameras in the spherical camera on the robot, respectively. $\Sigma_{X^r}$ denotes the robot coordinate at pose $X$.}
  \label{f:cood}
  \vspace*{-5mm}
\end{figure}

We define the robot and camera coordinates based on the global robot pose $X$, as shown in Fig.~\ref{f:cood}. $\Sigma_{X^r}$ is the base coordinate of the robot. In addition, $\Sigma_{X^f}$ and $\Sigma_{X^b}$ are the camera coordinates of the front- and back-side fisheye cameras in the spherical camera, respectively. 
{\colornewparts The axes directions are shown in Fig.~\ref{f:cood}. 
Following convention of the camera coordinate, we define the $y$ axis as downward and the $z$ axis as forward in $\Sigma_{X^f}$ and $\Sigma_{X^b}$.
Additionally, $\Sigma_{X^f}$ and $\Sigma_{X^b}$ are opposite around $y$ axis on their coordinates. 
We assume that the relative poses between each coordinate are known after measuring the height $h_{cam}$ and the offset $l_{cam}$ of the camera position. }
\subsection{Overview}
\subsubsection{Process of depth estimation}
Fig.~\ref{f:block}[a] presents the calculation of depth estimation for real images. Since there are no GT depth images, we employed self-supervised learning approach using time-sequence images. Unlike previous approaches\cite{zhou2017unsupervised,godard2019digging}, our spherical camera can capture both the front- and back-side of the robot. Hence, we propose a cost function to blend the front- and back-side images and thereby reduce the negative effects of occlusion. 

We feed the front-side images $I_{A^f}$ and back-side image $I_{A^b}$ at robot pose $A$ into the depth network $f_{\mbox{depth}}()$ to estimate the corresponding depth images as $D_{A^f}, D_{A^b} = f_{\mbox{depth}}(I_{A^f}, I_{A^b})$. 
By back-projection $f_{\mbox{backproj}}()$ with our proposed camera model, we can obtain the corresponding point clouds $Q_{A^f}$ and $Q_{A^b}$ for each camera coordinate $\Sigma_{A^f}$ and $\Sigma_{A^b}$, respectively.
\begin{eqnarray}
    Q_{A^f} = f_{\mbox{backproj}}(D_{A^f}), \hspace{2mm}Q_{A^b} = f_{\mbox{backproj}}(D_{A^b})
    \label{eq:bproj}
\end{eqnarray}
{\colornewparts Opposed to the baseline methods, our method predicts both the front- and back-side images from a single side image to blend them in $J_{bimg}$. }
Hence, ``Trans.'' in Fig.~\ref{f:pull}[a] transforms the coordinates of the estimated point clouds as follows:
\begin{eqnarray}
    && \hspace{-3mm}Q^{B^f}_{A^f} = T_{AB} Q_{A^f}, \hspace{8mm}Q^{B^f}_{A^b} = T_{AB} \cdot T^{-1}_{fb} Q_{A^b}, \\ \nonumber
    && \hspace{-3mm}Q^{B^b}_{A^f} = T_{fb} \cdot T_{AB} Q_{A^f}, \hspace{1mm}Q^{B^b}_{A^b} = T_{fb} \cdot T_{AB} \cdot T^{-1}_{fb} Q_{A^b}.
    \label{eq:trans}
\end{eqnarray}
Here, $Q^{Y^\beta}_{X^\alpha}$ denotes the point clouds on the coordinate $\Sigma_{Y^\beta}$ estimated from image $I_{X^\alpha}$. $T_{AB}$ is the estimated transformation matrix between $\Sigma_{A^f}$ and $\Sigma_{B^f}$. $T_{fb}$ is the known transformation matrix between $\Sigma_{X^f}$ and $\Sigma_{X^b}$. By projecting these point clouds with our learnable camera model, we can estimate four matrices $M^{B^\beta}_{A^\alpha}$, 
\begin{eqnarray}
    M^{B^\beta}_{A^\alpha} = f_{\mbox{proj}}(Q^{B^\beta}_{A^\alpha})
    \label{eq:proj}
\end{eqnarray}
where, $\alpha=\{f, b\}$ and $\beta=\{f, b\}$.
According to \cite{jaderberg2015spatial}, we estimate $I_{A^\alpha}$ by sampling the pixel value of $I_{B^\beta}$ as $\hat{I}^{B^\beta}_{A^\alpha} = f_{\mbox{sample}}(M^{B^\beta}_{A^\alpha}, I_{B^\beta})$. Here $\hat{I}^{B^\beta}_{A^\alpha}$ denotes the estimated image of $I_{A^\alpha}$ by sampling $I_{B^\beta}$. Note that we estimated four images from the combination of $\alpha=\{f, b\}$ and $\beta=\{f, b\}$, as shown in Fig.~\ref{f:block}[a]. We calculate the blended image loss $J_{bimg}$ to penalize the model during training.

\vspace{-3mm}
{\small
\begin{eqnarray}
    && \hspace{-7 mm} J_{bimg} = \lambda_{1} J_{L1} + \lambda_{2} J_{SSIM}. \\
    && \hspace{-7 mm} J_{L1} = \sum_{\alpha \in \{f, b\}} f_{\mbox{min}}(M|I_{A^\alpha}-\hat{I}^{B^f}_{A^\alpha}|, M|I_{A^x}-\hat{I}^{B^b}_{A^\alpha}|), \nonumber \\
    && \hspace{-7 mm} J_{SSIM} = \sum_{\alpha \in \{f, b\}} f_{\mbox{min}}(M d_{\mbox{ssim}}(I_{A^\alpha}, \hat{I}^{B^f}_{A^\alpha}), M d_{\mbox{ssim}}(I_{A^\alpha}, \hat{I}^{B^b}_{A^\alpha})). \nonumber
\end{eqnarray}
}
Here, $f_{\mbox{min}}(\cdot, \cdot)$ selects a smaller value at each pixel and calculates the mean of all the pixels. $d_{\mbox{ssim}}(\cdot, \cdot)$ is the pixel-wise structural similarity~(SSIM)~\cite{wang2004image}, following \cite{godard2019digging}. By selecting a smaller value in $f_{\mbox{min}}(\cdot, \cdot)$, we can equivalently select the non-occluded pixel value of $\hat{I}^{B^f}_{A^\alpha}$ or $\hat{I}^{B^b}_{A^\alpha}$ to calculate L1 and SSIM~\cite{godard2019digging}. $M$ is a mask to remove the pixels without RGB values and those of the robot itself, which are depicted in gray color in Fig.~\ref{f:pull}[a]. 
\subsubsection{Process of pose estimation}
Fig.~\ref{f:block}[c] denotes the process to estimate $T_{AB}$. Unlike previous monocular depth estimation approaches, we use the GT transformation matrix $\bar{T}_{AB}$ from the integral of the reference velocities $\{v_i, \omega_i \}_{i=0\cdots N_{g-1}}$ to move between poses $A$ and $B$. $J_{pose}$ is designed as $J_{pose} = \sum (\bar{T}_{AB} - T_{AB})^2$.
\begin{figure}[t]
  \vspace{1.5mm}
      \begin{center}
      \hspace{-8mm}\includegraphics[width=0.7\hsize]{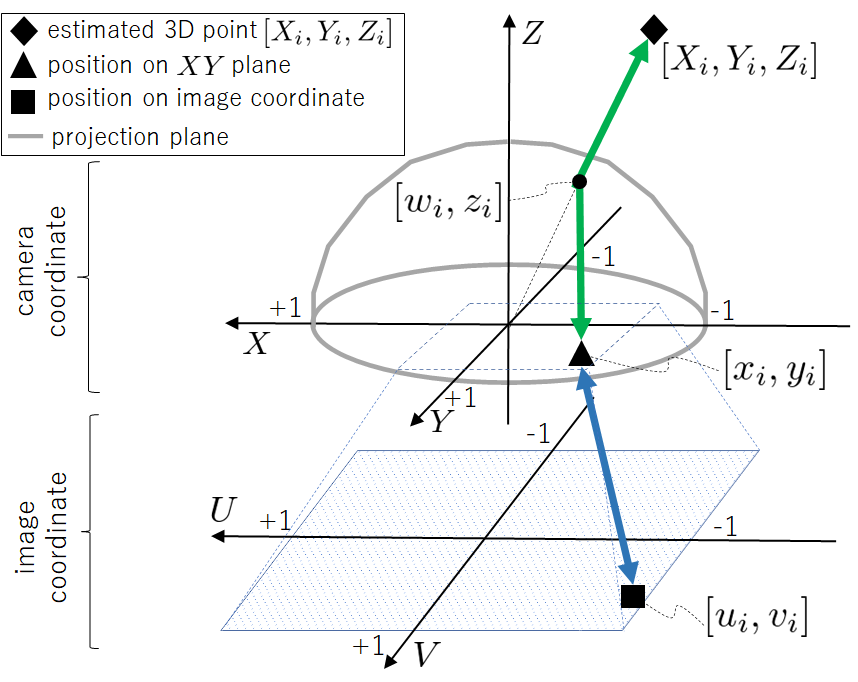}
      \end{center}
      \vspace*{-3mm}
	\caption{\small {\bf Overview of our camera model.} Our model projects and back-projects via $[x_i, y_i]$ on $XY$ plane of the camera coordinate.}
  \label{f:overview_cam}
  \vspace*{-3mm}
\end{figure}
\begin{figure}[t]
  \vspace{1.5mm}
  \begin{center}
      \includegraphics[width=0.9\hsize]{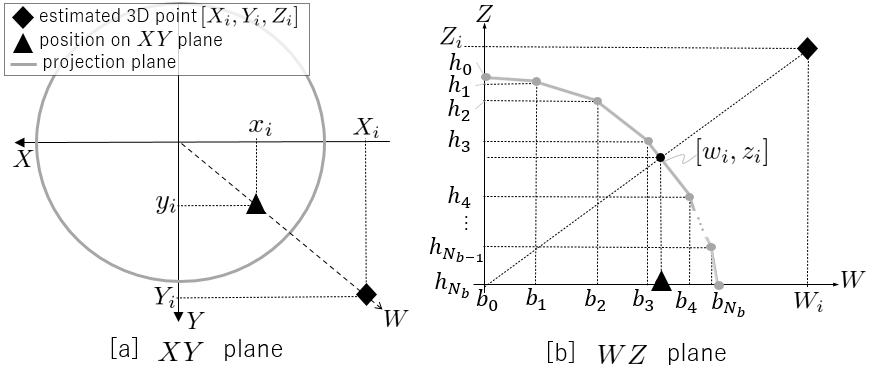}
  \end{center}
      \vspace*{-3mm}
	\caption{\small {\bf Projection surface of our learnable axisymmetric camera model.} Our projection surface using multiple line segments is axisymmetric and convex upward to being fully differentiable.}
  \label{f:cam}
  \vspace*{-5mm}
\end{figure}
\subsection{Learnable axisymmetric camera model}
{\colornewparts Our camera model defines the relationship between the pixel position $[u_i, v_i]$ on the image coordinates and corresponding 3D point $[X_i, Y_i, Z_i]$ on the camera coordinates. This mapping has been written as $f_{\mbox{backproj}}()$ or $f_{\mbox{proj}}()$ during the training process. Since we formulate a differentiable model, all parameters in our model can be simultaneously trained along with depth and pose networks. 

Figure~\ref{f:overview_cam} shows an overview of the camera model. There are two individual processes: 1) $[u_i, v_i] \Leftrightarrow [x_i, y_i]$~(blue double arrow) for modeling the field of view~(FoV) and the offset similar to the camera intrinsic parameters, and 2) $[x_i, y_i] \Leftrightarrow [X_i, Y_i, Z_i]$~(green double arrow) for handling the camera distortion. These processes are connected at $[x_i, y_i]$ on $XY$ plane of the camera coordinate. We explain each of the separate processes in the following paragraphs. To accept arbitrarily sized images, the image coordinate $UV$ is regularized within [-1, 1] and origin is positioned at the center of the image.}

\subsubsection{Part I~$[u_i, v_i] \Leftrightarrow [x_i, y_i]$}
{\colornewparts The mapping between $[x_i, y_i]$ and $[u_i, v_i]$ is defined by four independent parameters: $r_x, r_y$ for the FoV, and $o_x, o_y$ for the offset between $UV$ and $XY$. %camera's angle of view
Assuming linear relationships, this part of $f_{\mbox{backproj}}()$ is written as follows,
\begin{eqnarray}
    [x_i, y_i]^T = R \cdot [u_i, v_i]^T + [o_x, o_y]^T,
    \label{eq:camera_1}
\end{eqnarray}
where $R=\mbox{diag}(1/r_x, 1/r_y)$. 
%$r_x, r_y$ and $o_x, o_y$ are estimated between 0 and 1.0. 
Owing to the linearity and regularity, defining the inverse transformation for  $f_{\mbox{proj}}()$ as $R^{-1}\cdot([x_i, y_i]^T - [o_x, o_y]^T)$ is straightforward. }
\subsubsection{Part II~$[x_i, y_i] \Leftrightarrow [X_i, Y_i, Z_i]$}
{\colornewparts Next, we show the process of the green double arrow between $[x_i, y_i]$ and $[X_i, Y_i, Z_i]$ in Fig.~\ref{f:overview_cam} to model the distortion. 
At first, we design the learnable projection surface~(grey color lines in Fig.~\ref{f:overview_cam}).
Then, we explain the computation procedure in $f_{\mbox{backproj}}()$ and $f_{\mbox{proj}}()$.}
\paragraph{Projection surface}
{\colornewparts Figure~\ref{f:cam} indicates the details of our projection surface, which is axisymmetric around the $Z$ axis and convex upwards. Unlike the baseline camera model~\cite{usenko2018double,fang2021self}, our projection surface is modeled as a linear interpolation of a discrete surface to effectively train the camera model in self-supervised learning, inspired by Bhat et al.~\cite{bhat2021adabins}. Note that Bhat et al.~\cite{bhat2021adabins} discritizes the estimated depth itself and interpolates them in supervised learning architecture for explicit depth estimation, unlike our approach.

This projection surface can define a unique mapping between the angle of incident light and radial position of projected point on the $XY$ plane, which reflects the distortion property of the camera.
The projection surface on the $WZ$ plane is defined as consecutive line segments by the points $[b_i, h_i]_{i=0\cdots N_b}$ in Fig.~\ref{f:cam}[b].
Here, the $W$ axis is defined as the radial direction from the origin towards $[X_i, Y_i, 0]$ as shown in Fig.~\ref{f:cam}[a].
Because the parameters $[b_i, h_i]_{i=0\cdots N_b}$ are normalized to stabilize the training process, the projection surface on the $XY$ plane can be depicted as a unit circle centered at the origin, as shown in Fig.~\ref{f:cam}[a].
By giving the constraint $\Delta h_i/\Delta b_i > 0$ and $\Delta b_i > 0$, the convex upwards shape can be ensured. This constraint is indispensable to be fully differentiable model, as shown in {\it b) Back-projection} and {\it c) Projection}. Here $\Delta h_i = \tilde{h}_{i-1} - \tilde{h}_i$, $\Delta b_i = \tilde{b}_{i-1} - \tilde{b}_i$. $\tilde{x}$ indicates the variable $x$ before applying normalization. }

Our camera network $f_{\mbox{cam}}()$ estimates all parameters in our camera model as follows:
\begin{eqnarray}
    \{\Delta b_i, \Delta h_i/\Delta b_i\}_{i=1\cdots N_b}, r_x, r_y, o_x, o_y = f_{\mbox{cam}}(f_{img})
    \label{eq:camera}
\end{eqnarray}
where $f_{img}$ is the image features from the depth encoder. 
In $f_{\mbox{cam}}()$, we provide a sigmoid function at the last layer to achieve $\Delta h_i/\Delta b_i > 0$ and $\Delta b_i > 0$. 
By simple algebra with $\tilde{h}_{N_b} = 0.0$ and $\tilde{b}_{N_b} = 1.0$, we can obtain $[\tilde{b}_i, \tilde{h}_i]_{i=0\cdots N_b -1}$. 
By performing normalization to stabilize the training process, we can obtain $h_i = \tilde{h}_i/\sum_{k=0}^{N_b} \tilde{h}_k$ and $b_i = \tilde{b}_i/\sum_{k=0}^{N_b} \tilde{b}_k$. 
\paragraph{Back-projection}
To calculate $[X_i, Y_i, Z_i]$ from the estimated depth $Z_i$ at $[x_i, y_i]$, we first calculate $[w_i, z_i]$, which is the intersection point on the projection surface. The line of the $j$-th line segment of the projection surface on the $WZ$ plane can be expressed as: 
\begin{eqnarray}
    Z = \frac{h_{j-1} - h_j}{b_{j-1} - b_j} \cdot W + \frac{h_{j-1} b_j - h_j b_{j-1}}{b_{j-1} - b_j}=\alpha_j W + \beta_j. \label{eq:line_pc}
\end{eqnarray}
To achieve a fully differentiable process, we calculate the intersection points between all lines of the projection surface and the vertical line $W=w_i~(=\sqrt{x_i^2+y_i^2})$. Then, we select the minimum height at all intersections as $z_i$ because the projection surface is upwardly convex.
\begin{eqnarray}
    z_i = \mbox{min}(\{\alpha_j w_i + \beta_j \}_{j=1 \cdots N_b})
    \label{eq:min_back}
\end{eqnarray}
Here, the min function is a differentiable function. 
{\colornewparts Note that searching the corresponding line segment by element-wise comparison instead of the above calculation is not differentiable. }
Based on $z_i$, we can obtain $X_i = \frac{Z_i}{z_i}\cdot x_i$ and $Y_i = \frac{Z_i}{z_i}\cdot y_i$. 

\paragraph{Projection}
Similarly to back-projection, we first calculate the intersection point $[w_i, z_i]$ and derive $[x_i, y_i]$ from $[X_i, Y_i, Z_i]$. The line between the origin and $[X_i, Y_i, Z_i]$ can be expressed as $Z = \frac{Z_i}{W_i} \cdot W = \gamma_i \cdot W$. Much like the back-projection process, the minimum value at all intersections on the $Z$ axis can be $z_i$. Hence, $z_i$ and $w_i$ can be derived as follows:
\begin{eqnarray}
    z_i = \mbox{min} \left( \left\{\frac{\gamma_i \beta_j}{\gamma_i - \alpha_j} \right\}_{j=1 \cdots N_b} \right), \hspace{1 mm}w_i = z_i / \gamma_i.
    \label{eq:min_proj}
\end{eqnarray}
Thus, $[x_i, y_i]$$=$$[\frac{w_i}{W_i}\cdot X_i, \frac{w_i}{W_i}\cdot Y_i]$. Here, $W_i$$=$$\sqrt{X_i^2 + Y_i^2}$.

\vspace{2mm}
{\colornewparts In both back-projection and projection, we clamp $z_i$ between $h_0$ and $h_{N_b}$, and do not consider the points $[x_i, y_i]$ outside the unit circle on the $XY$ plane as the out of view points. }
\subsection{Floor loss}
The highly reflective floor surface in indoor environments can often cause significant artifacts in depth estimation. Our floor loss function provides geometric supervision for floor areas. Assuming that the camera is horizontally mounted on the robot and its height is known, the floor loss function is constructed by two components: $J_{floor} = J_{fcl} + J_{lbl}$.
\subsubsection{Footprint consistency loss $J_{fcl}$}
The robot footprint is almost horizontally flat, and its height is obtained from the camera mounting position. According to the method shown below, we obtain the GT depth $\{ \bar{D}_{A^\alpha} \}_{\alpha = \{f, b\}}$ only around the robot footprint between $\pm M_r$ steps and provide the supervision as follows: 
\begin{eqnarray}
    J_{fcl} = \sum_{\alpha \in \{f, b\}} \frac{1}{N_m}\sum_{i=1}^{N_m} M_f |\bar{D}_{A^\alpha} - D_{A^\alpha}|,
    \label{eq:foc}
\end{eqnarray}
where $M_f$ is the mask, which masks out the pixels without the GT values in $\bar{D}_{A^\alpha}$, and $N_m$ is the number of pixel with the GT depth. $\bar{D}_{A^\alpha}$ can be derived as:
\begin{eqnarray}
    \bar{D}_{A^\alpha}(x^{\alpha}_j, y^{\alpha}_j) = Z^{\alpha}_j, 
    \label{eq:dabar}
\end{eqnarray}
where $[x^{\alpha}_j, y^{\alpha}_j] = f_{\mbox{proj}}(T_{r\alpha}Q_{foot}[j])$ and $ Z^{\alpha}_j$ are the values of the $Z$ axis of $T_{r\alpha}Q_{foot}[j]$. Here, $Q_{foot}$ is the point clouds of the robot footprint between $\pm M_r$ steps on $\Sigma_{X^r}$. {\colornewparts To obtain $Q_{foot}$, we calculate the robot local positions using the teleoperator's velocity between $\pm M_r$ steps.} And, $T_{r\alpha}$ is the known transformation matrix from $\Sigma_{X^r}$ to $\Sigma_{X^\alpha}$. 
By assigning all point clouds into $\bar{D}_{A^\alpha}$ using (\ref{eq:dabar}), we can take $\bar{D}_{A^\alpha}$ to calculate $J_{fcl}$. Note that $\bar{D}_{A^\alpha}$ is a sparse matrix. No GT pixel in $\bar{D}_{A^\alpha}$ is excluded by $M_f$. 
\subsubsection{lower boundary loss $J_{lbl}$}
In indoor scenes, floor areas often reflect ceiling lights. This can cause it to appear that there are holes on the floor in the estimated depth image. To provide a lower boundary for the height of estimated point clouds, we observe two key points: 1) the camera is horizontally mounted on the robot, and 2) most objects around the robot are higher than the floor. One exception would be anything that is downstairs, which would be lower than the floor. However, it is rare in to find such occurrences in the dataset, because teleoperation around the stairs is dangerous and ill-advised.

To provide the constraint for the $Y$ axis~($=$ hight) value of the estimated point clouds, $J_{lbl}$ can be given as follows: 
\begin{eqnarray}
    J_{lbl} = \sum_{\alpha \in \{f, b\}} \frac{1}{N}\sum_{i=1}^N(\mbox{max}(0.0, Y_{A^\alpha} - h_{cam})),
    \label{eq:flc}
\end{eqnarray}
where $Q_{A^\alpha} = [X_{A^\alpha}, Y_{A^\alpha}, Z_{A^\alpha}]$. $J_{lbl}$ penalizes $Y_{A^\alpha}$, which is larger~($=$ lower) than the floor height ($=h_{cam}$). 
\subsection{Sim2real transfer with $J_{depth}$}
In conjunction with self-supervised learning using time-sequence real images, we used the GT depth image $\{\bar{D}_{s^f}, \bar{D}_{s^b}\}$ from a photo-realistic robot simulator~\cite{xia2020interactive,igibson2}, as shown in Fig.~\ref{f:block}[b]. Although there is an appearance gap between real and simulated images, the GT depth from the simulator can help the model understand the 3D geometry of the environment from the image. 
Here, $J_{depth} = \sum_{\alpha \in \{f, b\}} d_{\mbox{depth}}(\bar{D}_{s^\alpha}, D_{s^\alpha})$, where $D_{s^f}, D_{s^b} = f_{\mbox{depth}}(I_{s^f}, I_{s^b})$. $I_{s^f}$ and $I_{s^b}$ are front- and back-side fisheye images, respectively, from the simulator. We employed the same metric $d_{\mbox{depth}}()$ as the baseline method~\cite{alhashim2018high} to measure the depth differences. To achieve a sim2real transfer, we simultaneously penalized $J_{depth}$ and $J_{bimg}$.
\begin{figure}[t]
  \vspace{1.5mm}
  \begin{center}
      \includegraphics[width=0.9\hsize]{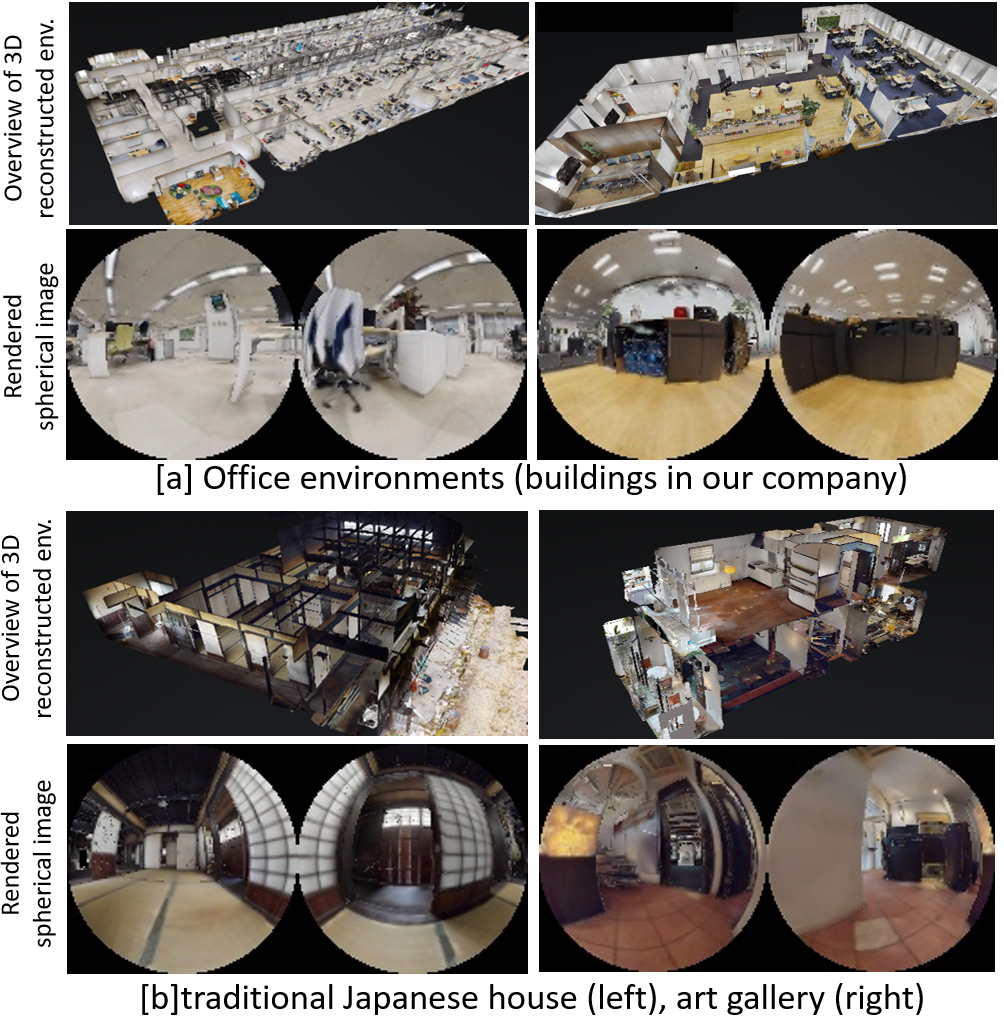}
  \end{center}
      \vspace*{-3mm}
	\caption{\small {\bf Examples of measured 3D reconstructed environments~\cite{matterport} and rendered spherical images~\cite{xia2020interactive,igibson2}.} [a] The office and laboratory from our company's buildings for training and validation, [b] traditional Japanese house and art gallery for testing. }
  \label{f:vchuken}
  \vspace*{-3mm}
\end{figure}
\section{EXPERIMENT}
\subsection{Dataset}
Our method was mainly evaluated using the GO Stanford~(GS) dataset with time-sequence spherical camera images.
In addition, we used the KITTI dataset with pinhole camera images for comparison with the baseline methods, which attempts to learn the camera parameters.
\subsubsection{GO Stanford dataset}
We used the GS dataset~\cite{hirose2018gonet,hirose2019vunet,hirose2019deep} with time-sequence spherical camera images~(256$\times$128) and the reference velocities, which were collected by teleoperating turtlebot2 with Ricoh THETA S.
The GS dataset contains 10.3 hours of data from twelve buildings at the Stanford University campus.
To train our networks, we used a training dataset from eight buildings, following \cite{hirose2019deep}.

In addition to the real images of the GS dataset, we collected pairs of simulator images and GT depth images for $J_{depth}$.
We scanned 12 floors (e.g., office rooms, meeting rooms, laboratories) in our company buildings and ten environments~(e.g., a traditional Japanese house, art gallery, fitness gym) for the simulator by Matterport pro2, as shown in Fig.~\ref{f:vchuken}. 
We separated them into groups: eight floors in our company buildings for training, four floors in our company buildings for validation, and ten environments not in our company for testing. 
For training, we rendered 10,000 data from each environment using interactive Gibson~\cite{xia2020interactive,igibson2}. In addition, we collected 1,000 data points for testing from ten environments. 

Data collection of the GT depth from a real spherical image is a challenging task. Hence, in the quantitative analysis, we used the GT depth from the simulator. To evaluate the generalization performance, our test environment was not derived from our company building. Examples of the domain gaps between training and testing are shown in Fig.~\ref{f:vchuken}. In the qualitative analysis, we used both real and simulated images.
\begin{table}[t]
  \vspace{1.5mm}
  \caption{{\small {\bf Evaluation of depth estimation from the GO Stanford dataset.} For the three leftmost metrics, smaller values are better; for the rightmost metric, higher values are better. ``SGT'' denotes the GT depth from the simulator. The bold values indicate the best results. All methods except $\dagger$ were evaluated without median scaling because scaling is learned from the SGT. $\ddagger$ used the half-sphere camera model shown in (\ref{eq:XY_app}) and (\ref{eq:xy_app}).}}%depth provided by the simulator
  \vspace*{-3mm}
  \begin{center}
  \resizebox{1.0\columnwidth}{!}{
  \label{tab:ev_gs}
  \begin{tabular}{lc|c|c|c|c} \hline
    Method & SGT & \cellcolor[rgb]{0.9, 0.5, 0.5}Abs-Rel & \cellcolor[rgb]{0.9, 0.5, 0.5}Sq-Rel & \cellcolor[rgb]{0.9, 0.5, 0.5}RMSE & \cellcolor[rgb]{0.5, 0.5, 0.9}$\delta$$<$$1.25$ \\ \hline
    monodepth2~\cite{godard2019digging}~$\ddagger$ $\dagger$ & & 0.586 & 1.890 & 1.123 & 0.439 \\
    \hspace{0mm}~--with SGT\cite{godard2019digging,alhashim2018high}~$\ddagger$& \checkmark & 0.228 & 0.162 & 0.535 & 0.664 \\
    Alhashim et al.~\cite{alhashim2018high} & \checkmark & 0.203 & 0.154 & 0.542 & 0.698 \\ \hdashline
    \bf{Our method~(full)} & \checkmark & \bf{0.198} & \bf{0.143} & \bf{0.525} & \bf{0.711} \\
    \hspace{0mm}~--wo $J_{depth}$~$\dagger$ & & 0.377 & 0.335 & 0.829 & 0.433 \\    
    \hspace{0mm}~--wo our cam. model~$\ddagger$ & \checkmark & 0.248 & 0.179 & 0.548 & 0.646 \\ 
    \hspace{0mm}~--wo $J_{lbl}$ & \checkmark & 0.233 & 0.161 & 0.532 & 0.665 \\    
    \hspace{0mm}~--wo $J_{fcl}$ & \checkmark & 0.216 & 0.151 & 0.530 & 0.684 \\
    \hspace{0mm}~--wo blending in $J_{bimg}$ & \checkmark & 0.205 & 0.147 & 0.530 & 0.702 \\   
    \hspace{0mm}~--wo $J_{pose}$ & \checkmark & 0.200 & 0.145 & 0.532 & 0.698 \\ \hline    
  \end{tabular}
  }
\end{center}
\vspace{-3mm}
\end{table}
\begin{table}[t]
  \caption{{\small {\bf Evaluation of depth estimation on the KITTI raw dataset.} For the leftmost three metrics, smaller values are better; for the rightmost metric, higher values are better. The bold values indicate the best results. All the methods in this table employ a pretrained ResNet-18 for their encoder. All values were calculated after median scaling. $\dagger$ was trained on a mixed dataset with the KITTI, Cityscape, bike, and GO Stanford datasets. The others were trained only on KITTI.}}
  \vspace*{-3mm}
  \begin{center}
  \resizebox{1.0\columnwidth}{!}{
  \label{tab:ev_kitti}
  \begin{tabular}{lc|c|c|c|c} \hline
    Method & Camera & \cellcolor[rgb]{0.9, 0.5, 0.5}Abs-Rel & \cellcolor[rgb]{0.9, 0.5, 0.5}Sq-Rel & \cellcolor[rgb]{0.9, 0.5, 0.5}RMSE & \cellcolor[rgb]{0.5, 0.5, 0.9}$\delta$$<$$1.25$\\ \hline
    Gordon et al.~\cite{gordon2019depth} & known & 0.129 & 0.982 & 5.23 & 0.840 \\
    Gordon et al.~\cite{gordon2019depth} & learned & 0.128 & 0.959 & 5.23 & 0.845 \\ \hdashline
    %Gordon et al.~\cite{gordon2019depth} & K+CS & learned & 0.124 & 0.930 & 5.12 & 0.851 \\ 
    NRS~\cite{9320335} & known & 0.137 & 0.987 & 5.337 & 0.830 \\
    NRS~\cite{9320335} & learned & 0.134 & 0.952 & 5.264 & 0.832 \\ \hdashline
    monodepth2~(original) & known & 0.115 & 0.903 & 4.864 & 0.877 \\
    \hspace{1mm}\bf{with our cam. model} & known & 0.115 & 0.915 & 4.848 & \bf{0.878} \\
    \hspace{1mm}\bf{with our cam. model} & learned & \bf{0.113} & \bf{0.885} & \bf{4.829} & \bf{0.878} \\  
    \hspace{1mm}\bf{with our cam. model}$\dagger$\cellcolor[rgb]{0.9, 0.9, 0.9} & \cellcolor[rgb]{0.9, 0.9, 0.9}learned & \bf{0.109}\cellcolor[rgb]{0.9, 0.9, 0.9} & \bf{0.826}\cellcolor[rgb]{0.9, 0.9, 0.9} & \bf{4.702}\cellcolor[rgb]{0.9, 0.9, 0.9} & \bf{0.884}\cellcolor[rgb]{0.9, 0.9, 0.9} \\ \hline           
    %\hspace{1mm}\bf{with our cam. model}$\dagger$\cellcolor[rgb]{0.9, 0.9, 0.9} & \cellcolor[rgb]{0.9, 0.9, 0.9}learned & \bf{0.111}\cellcolor[rgb]{0.9, 0.9, 0.9} & \bf{0.811}\cellcolor[rgb]{0.9, 0.9, 0.9} & \bf{4.631}\cellcolor[rgb]{0.9, 0.9, 0.9} & \bf{0.881}\cellcolor[rgb]{0.9, 0.9, 0.9} \\ \hline       
  \end{tabular}
  }
\end{center}
\vspace{-5mm}
\end{table}
\subsubsection{KITTI dataset}
To evaluate our proposed camera model against the baseline methods, we employed the KITTI raw dataset~\cite{Geiger2013IJRR} {\colornewparts for the evaluation of depth estimation}.
Similar to the baseline methods, we separated the KITTI raw dataset via Eigen split~\cite{eigen2014depth} with 40,000 images for training, 4,000 images for validation, and 697 images for testing. 
To compare with baseline methods, we employed the widely used 640$\times$192 image size as input.

{\colornewparts Besides, we used the KITTI odometry dataset with ground truth pose for the evaluation of pose estimation. It is known that the KITTI raw dataset for depth estimation partially includes test images from the KITTI odometry dataset. Hence, following the baseline methods, we trained our models with sequences 00 to 08 and conduct testing on sequences 09 and 10.}
\begin{figure*}[ht]
  \vspace{1mm}
  \begin{center}
      \includegraphics[width=0.95\hsize]{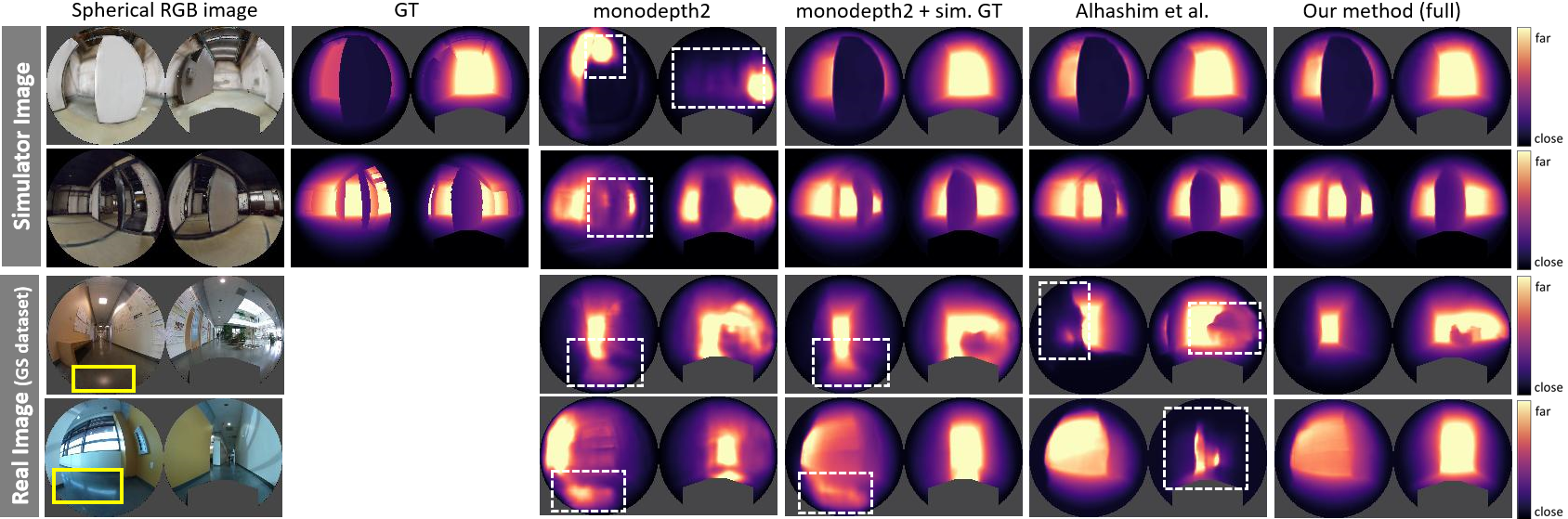}
  \end{center}
      \vspace*{-3mm}
	\caption{\small {\bf Estimated depth images from 256$\times$128 spherical camera image~(GS dataset) provided by the baseline methods and our method~(full).} The top two rows represent the simulator image and the bottom two rows represent the real image. Yellow rectangles in the RGB image highlight the floor surface with reflected ceiling lights. White dashed rectangles on depth estimations highlight the artifacts.}
  \label{f:gs}
  \vspace*{-5mm}
\end{figure*}
\subsection{Training}
In training with the GS dataset, we randomly selected 12 real images from the training dataset as $\{I_{A^\alpha}\}_{\alpha \in \{f, b\}}$. Then, we randomly selected $\{I_{B^\alpha}\}_{\alpha \in \{f, b\}}$ between $\pm$5(= $N_g$) steps. In addition, we selected 12 simulator images $\{I_{s^\alpha}\}_{\alpha \in \{f, b\}}$ and the GT depth $\{\bar{D}_{s^\alpha}\}_{\alpha \in \{f, b\}}$ for $J_{\mbox{depth}}$. 

{\colornewparts To define the transformation matrices $T_{fb}$, $T_{rf}$, and $T_{rb}$, we measured $h_{cam}$ and $l_{cam}$ in Fig.~\ref{f:cood} as 0.57 and 0.12 meters, respectively. Additionally, we set $N_b=32$ for our camera model.}
The weighting factors for the loss function $J$ were designed as $\lambda_{1} = 0.85$, $\lambda_{2} = 0.15$, $\lambda_{s} = 0.001$, $\lambda_{f} = 0.1$, and $\lambda_{d} = \lambda_{p} = 1.0$. $\lambda_{1}$, $\lambda_{2}$, and $\lambda_{s}$ were exactly the same as in previous studies. We only determined $\lambda_{f}$ by trial and error.

The robot footprint shape was defined as a circle with a diameter of 0.5 m. The point cloud of the footprint was set as 1400 points inside the circle. The number of steps for the robot footprints was $M_r=5$ for $J_{fcl}$.

The network structures of $f_{\mbox{depth}}()$ and $f_{\mbox{pose}}()$ were exactly the same as those of monodepth2~\cite{godard2019digging}. In addition, $f_{\mbox{cam}}()$ was designed with three convolutional layers, with the ReLu function and two fully connected layers with a sigmoid function, to estimate the camera parameters.
We used the Adam optimizer with a learning rate of 0.0001 and conducted training loop for 40 epochs.

During training, we iteratively calculated $J$ and derived the gradient to update all the models. Hence, we could simultaneously penalize $J_{\mbox{depth}}$ from the simulator and the others from the real images to achieve a sim2real transfer.

For the KITTI dataset, we employed the source code of monodepth2~\cite{godard2019digging} and replaced the camera model with our proposed camera model. The other settings were exactly the same as those of monodepth2, to focus the experimentation on our camera model.
\subsection{Evaluation of depth estimation}
\subsubsection{Quantitative Analysis}
Table~\ref{tab:ev_gs} shows the ablation study of our method and the results of three baseline methods for comparison. We trained the following baseline methods with the same dataset. 

\vspace{1mm}
\noindent
\textbf{monodepth2~\cite{godard2019digging}} We applied the following half-sphere model~\cite{courbon2007generic} into monodepth2~\cite{godard2019digging}, instead of the pinhole camera model, and trained depth and pose networks. {\colornewparts This model assumes that the $UV$ coordinate is the same as the $XY$ coordinate and the projection surface is a half-sphere. }%using GS dataset.

\vspace{2mm}
\noindent
$\bullet$ back-projection: ($u_i$, $v_i$, $Z_i$) $\rightarrow$ ($X_i$, $Y_i$),
\begin{eqnarray}
    X_i = \frac{Z_i}{\sqrt{1 - u_i^2 -v_i^2}} \cdot x_i, \hspace{2mm}
    Y_i = \frac{Z_i}{\sqrt{1 - u_i^2 -v_i^2}} \cdot y_i
    \label{eq:XY_app}
\end{eqnarray}

\noindent
$\bullet$ Projection: ($X_i$, $Y_i$, $Z_i$) $\rightarrow$ ($u_i$, $v_i$)
\begin{eqnarray}
    u_i = \frac{X_i}{\sqrt{X_i^2 + Y_i^2 + Z_i^2}}, \hspace{2mm}
    v_i = \frac{Y_i}{\sqrt{X_i^2 + Y_i^2 + Z_i^2}}
    \label{eq:xy_app}
\end{eqnarray}

\vspace{1mm}
\noindent
\textbf{monodepth2 with sim. GT~\cite{godard2019digging, alhashim2018high}} We added $J_{depth}$  to the cost function of the above baseline to train the models.

\vspace{1mm}
\noindent
\textbf{Alhashim et al.~\cite{alhashim2018high}} 
We trained the depth network by minimizing $J_{depth}$, which is the same cost function as \cite{alhashim2018high}. 
\vspace{0mm}

{\colornewparts
In quantitative analysis, we evaluate the estimated depth using common metrics.
``Abs-Rel,'' ``Sq-Rel,'' and ``RMSE,'' are calculated by means of the following values. 
\begin{itemize}
  \item Abs-Rel\hspace{9mm}:\hspace{2mm} $|D_{gt} - \hat{D}|/D_{gt}$
  \item Sq Rel\hspace{11mm}:\hspace{2mm} $(D_{gt} - \hat{D})^2/D_{gt}$
  \item RMSE\hspace{11mm}:\hspace{2mm} $((D_{gt} - \hat{D})^2)^{\frac{1}{2}}$
\end{itemize}
Here, $D_{gt}$ is the ground truth of the estimated depth image $\hat{D}$. The remaining metric is the ratios that satisfy $\delta < 1.25$. $\delta$ is defined as $\delta = \mbox{max}(D_{gt}/\hat{D}, \hat{D}/D_{gt})$.
}

From Table~\ref{tab:ev_gs}, we can observe that our method significantly outperforms all baseline methods. In addition, we confirmed the advantages of our proposed components via the ablation study. The use of the GT depth from the simulator and the learning axisymmetric camera model were fairly effective. 
{\colornewparts Even though the proposed method is evaluated without scaling using the GT depth, it outperforms the other methods with scaling. This suggests that our method learns the correct scaling via $J_{depth}$ with the GT depth from the simulator. } 

In Table~\ref{tab:ev_kitti}, we present the quantitative results for the KITTI dataset. Similar to our method, the baseline methods (shown in Table~\ref{tab:ev_kitti}) learned the camera model. All methods presented in Table~\ref{tab:ev_kitti} used ResNet-18 for their depth network to allow for fair comparisons. 

In our method with known camera parameters, we set $\{h_q \}_{q=0 \cdots N_b}$, $r_x$, $r_y$, $o_x$, and $o_y$ as the constant values for the GT camera’s intrinsic parameters. $\{b_q \}_{q=0 \cdots N_b}$ was designed with equal intervals between 0.0 and 1.0.
Our method improved the accuracy of depth estimation by learning the camera model. In addition, our method outperformed all baseline methods, including the original monodepth2. 

Moreover, we trained our models by mixing the KITTI, Cityscape~\cite{Cordts2016Cityscapes}, bike~\cite{mahjourian2018unsupervised}, and GS datasets~\cite{hirose2019deep} to evaluate its ability to handle various cameras and evaluated on KITTI images. During training, all images were aligned into KITTI's image size by center cropping. In GS dataset, we use front-side fisheye images. Our method showed improved performance by adding datasets from various cameras with various distortions, as shown at the bottom of Table~\ref{tab:ev_kitti}. % without specific fine-tuning on KITTI. , which include fisheye cameras and iPhones
\subsubsection{Qualitative Analysis}
Figure~\ref{f:gs} shows the estimated depth images from simulated images and real images from the GS dataset.
The depth images estimated by {\bf monodepth2} are blurred. This is caused by the small size of the input image and the camera model’s error.
The GT depth from the simulator can sharpen the simulated depth of images in {\bf monodepth2 with sim. GT}. However, there are many artifacts, particularly on the reflected floor.
{\bf Alhashim et al.} observed errors in the estimated depth from real images. {\bf Alhashim et al.} failed sim2real transfer because the depth network was trained only from simulated images. However, our method~(rightmost side) can accurately estimate depth images of both real and simulated images by reducing these artifacts. Additional examples are provided in the supplemental videos.

Finally, we present the depth images of KITTI in Fig.~\ref{f:pull} [b]. Our method can handle the pinhole camera image and estimate an accurate depth image without camera calibration.
{\colornewparts
\subsection{Evaluation of pose estimation}
We evaluate our pose network using the KITTI odometry dataset with the ground truth poses. Table~\ref{tab:pose_kitti} shows the mean and standard deviation of absolute trajectory error over five-snippets in the test dataset, following the baseline methods. Although Gordon et al.~\cite{gordon2019depth} with known camera model shows explicit advantages, their performance deteriorates while learning the camera parameters. Besides, our method with learning our camera model shows a healthy advantageous gap against the original monodepth2 with known camera intrinsic parameters. 
}
\begin{table}[t]
  \caption{{\small {\bf Evaluation of pose estimation on the KITTI odometry dataset.} Mean and standard deviation of absolute trajectory error~(ATE) over five-frame snippets are calculated for sequence 09 and 10, respectively. The bold value indicates the better one between "known" or  "learned".}}
  \vspace*{-3mm}
  \begin{center}
  \resizebox{1.0\columnwidth}{!}{
  \label{tab:pose_kitti}
  \begin{tabular}{lc|c|c} \hline
    Method & Camera & Sequence 09 & Sequence 10 \\ \hline
    Gordon et al.~\cite{gordon2019depth} & known & \bf{0.009  $\pm$ 0.0015} & \bf{0.008 $\pm$ 0.011} \\
    Gordon et al.~\cite{gordon2019depth} & learned & 0.0120  $\pm$ 0.0016 & 0.010  $\pm$ 0.010 \\ \hdashline
    NRS~\cite{9320335} & known & -- & -- \\
    NRS~\cite{9320335} & learned & 0.0150 $\pm$ 0.0301 & 0.0103 $\pm$ 0.0073 \\ \hdashline
    monodepth2~(original) & known & 0.017  $\pm$ 0.008  & 0.015  $\pm$ 0.010  \\
    \hspace{1mm} {\bf with our cam. model} & learned & \bf{0.0134 $\pm$ 0.0068} & \bf{0.0134 $\pm$ 0.0084} \\ \hline
  \end{tabular}
  }
\end{center}
\vspace{-3mm}
\end{table}
\section{CONCLUSIONS}
We proposed a novel learnable axisymmetric camera model for self-supervised monocular depth estimation. In addition, we proposed to supervise the estimated depth using the GT depth from the photo-realistic simulator. By mixing real and simulator images during training, we can achieve a sim2real transfer in depth estimation. Additionally, we proposed loss functions to provide the constraints for the floor to reduce artifacts that may result from reflective floors. The effectiveness of our method was quantitatively and qualitatively validated using the GS and KITTI datasets. 
\section{ACKNOWLEDGMENT}
We thank Kazutoshi Sukigara, Kota Sato, Yuichiro Matsuda, and Yasuaki Tsurumi for measuring 3D environments to collect pairs of simulator images and GT depth images. 
%
%\clearpage
%\balance
%\bibliographystyle{unsrt}
\bibliographystyle{IEEEtran}
\vskip-\parskip
\begingroup
\footnotesize
\bibliography{egbib}
\endgroup

\end{document}